\documentclass[twoside,twocolumn,9pt]{article}
\usepackage{extsizes}
\usepackage[super,sort&compress,comma]{natbib} 
\usepackage[version=3]{mhchem}
\usepackage[left=1.5cm, right=1.5cm, top=1.785cm, bottom=2.0cm]{geometry}
\usepackage{balance}
\usepackage{mathptmx}
\usepackage{sectsty}
\usepackage{graphicx} 
\usepackage{lastpage}
\usepackage[format=plain,justification=justified,singlelinecheck=false,font={stretch=1.125,small,sf},labelfont=bf,labelsep=space]{caption}
\usepackage{amsfonts}
\usepackage{float}
\usepackage{fancyhdr}
\usepackage{fnpos}
\usepackage[english]{babel}
\addto{\captionsenglish}{%
  
}
\usepackage{array}
\usepackage{droidsans}
\usepackage{charter}
\usepackage[T1]{fontenc}
\usepackage[usenames,dvipsnames]{xcolor}
\usepackage{setspace}
\usepackage[compact]{titlesec}
\usepackage{hyperref}
\usepackage{booktabs}
\usepackage{amsmath}
\usepackage{epstopdf}

\definecolor{cream}{RGB}{222,217,201}
\newcommand{\CorpusCandidates}{2,777}  % Scopus hits before any filter. The top of the funnel.
\newcommand{\CorpusFiltered}{1,575}  % Papers passing the title and abstract filter -- how many relevant papers exist. The denominator for every coverage claim.
\newcommand{\CorpusConverted}{1,027}  % Papers actually obtained and turned into chunked text. Below CorpusFiltered because of access, not relevance.
\newcommand{\DatabasePapers}{1,026}  % Papers the extractor processed.
\newcommand{\DatabaseYielding}{664}  % Papers that produced at least one record.
\newcommand{\DatabaseEmpty}{362}  % Papers that produced none. Mostly reviews and enzymatic work, i.e. papers that should yield none -- but this is also the audit's blind spot, since a missed experiment is invisible.
\newcommand{\DatabaseRecords}{5,563}  % Records extracted across the whole corpus.
\newcommand{\EmptyUnclassified}{202}  % Papers that yielded no records and whose titles did not say why. The residual of the three named reasons.
\newcommand{\EmptyBiological}{92}  % Papers yielding nothing because they concern enzymatic or biological degradation, which is out of scope.
\newcommand{\EmptyOffTarget}{57}  % Papers yielding nothing because they follow another route or use PET only as a material.
\newcommand{\EmptyReview}{11}  % Papers yielding nothing because they are reviews without protocols.
\newcommand{\CitationsTotal}{14,098}  % Chunk citations across the database.
\newcommand{\CitationsResolved}{13,602}  % Citations pointing at a chunk that exists in the paper the record came from.
\newcommand{\CitationsTraceable}{96.5}  % Resolved as a share of all citations.
\newcommand{\IdentityPairs}{1,423}  % Records reporting both yield and conversion, so the identity can be tested.
\newcommand{\IdentityImpossible}{28}  % Of those, records with yield above conversion. An identity, not a correlation.
\newcommand{\RouteGlycolysisShare}{48}  % Glycolysis records as a percentage of the database.
\newcommand{\RouteHydrolysisShare}{25}  % Hydrolysis records as a percentage of the database.
\newcommand{\RouteMethanolysisShare}{10}  % Methanolysis records as a percentage of the database.
\newcommand{\CuratedExperiments}{295}  % Experiments in the hand-curated answer key, including the entries the supervisors confirmed in the rescue round. Every metric-grader score is computed against this.
\newcommand{\CuratedPapers}{24}  % Papers the curated table covers.
\newcommand{\AdjRecords}{112}  % Records the chemists ruled on.
\newcommand{\AdjPairs}{51}  % Discordant pairs: records where exactly one grader matched the chemists. The only records that say which grader is better.
\newcommand{\AdjFavourJudge}{43}  % Discordant pairs won by the judge.
\newcommand{\AdjFavourMetric}{8}  % Discordant pairs won by the metric.
\newcommand{\AdjMcNemarP}{7\times10^{-7}}  % McNemar exact p on those pairs. Every pair is censused, so this carries no sampling error.
\newcommand{\AdjJudgeAgrees}{85.7}  % How often the judge's verdict matches the chemists', over the reviewed records. Not AdjJudgeWinRate, which is its share of the records the two graders disagree on: that is a different question over a different denominator, and the abstract quoted it for this one.
\newcommand{\ShotsDelta}{+0.025}  % F1 gain from one worked example over none.
\newcommand{\ShotsP}{0.082}  % p for that gain.
\newcommand{\SourceDelta}{+0.064}  % F1 gain from requiring source citations.
\newcommand{\SourceCiting}{0.780}  % F1 with citations required.
\newcommand{\SourceCitingSd}{0.021}  % Its spread.
\newcommand{\SourceNot}{0.716}  % F1 without.
\newcommand{\SourceNotSd}{0.043}  % Its spread.
\newcommand{\SourceP}{0.079}  % p for the difference. Short of significance, reported as such.
\newcommand{\CitationsUntraceable}{3.5}  % Unresolved as a share -- the complement of CitationsTraceable.
\newcommand{\RouteOther}{922}  % Records whose solvent names no route, so no product is fixed.
\newcommand{\RouteNamedTotal}{4,641}  % The three named routes summed. With RouteOther it accounts for the corpus.
\newcommand{\MatrixRecords}{296}  % Benchmark records each judge-extractor pair was scored over. The denominator of the agreement figures and of the agreement pie in fig_graders.
\newcommand{\AdjAcceptedReviewed}{49}  % Records in the both-accepted cell under today's verdicts.
\newcommand{\JudgeContextCap}{131{,}072}  % The judge model's context window, in tokens.
\newcommand{\FunnelDroppedPolymer}{550}  % Dropped: no polymer named.
\newcommand{\FunnelDroppedRoute}{381}  % Dropped: no depolymerisation route named.
\newcommand{\FunnelDroppedOffTopic}{271}  % Dropped: off-topic subject or a review.
\newcommand{\FunnelPassRate}{56.7}  % Share of candidates surviving the filter.
\newcommand{\DatabaseCost}{8.63}  % USD spent on the mass extraction.
\newcommand{\GaoPapers}{19}  % Papers those came from, identified from Gao's short labels by hand.
\newcommand{\WithinPaperRho}{0.34}  % Median within-paper Spearman rho for temperature against yield. Positive, i.e. the expected chemistry, once each paper is compared only with itself.
\newcommand{\WithinPaperP}{7\times10^{-7}}  % Wilcoxon signed-rank p for those within-paper correlations against zero. Was typed into the body by hand as 7e-7 while the check had been returning it all along.
\newcommand{\WithinPooledRho}{+0.018}  % The same correlation pooled across every paper: near zero. Pooling is the wrong operation on optimised experiments, and the contrast with WithinPaperRho is the point.
\newcommand{\WithinPaperPositive}{85}  % Share of those papers where the correlation runs the predicted way.
\newcommand{\AdjBothFlagged}{12}  % Records the page censused because both graders flagged them.
\newcommand{\CorpusNotObtained}{548}  % Filtered minus converted: papers we know exist and cannot read. Concentrated in ACS and RSC.
\newcommand{\CorpusElsevierXml}{447}  % Obtained as Elsevier full-text XML.
\newcommand{\CorpusPdf}{488}  % Obtained as PDF: Wiley and Springer under entitlement, the rest open access.
\newcommand{\CorpusEuropePmc}{89}  % Obtained as Europe PMC JATS deposits.
\newcommand{\CorpusOtherSource}{3}  % Obtained by no other named route.
\newcommand{\ThresholdAccept}{0.30}  % The accept threshold: a matched pair counts as a true positive below this mean record penalty. Frozen at 0.30 and read straight from checks/_setup.py.
\newcommand{\ThresholdCatalyst}{0.60}  % The catalyst gate: two catalyst names must be at least this similar (SequenceMatcher ratio) for a pair to match at all. The Methodology stated 0.80; the value used to compute every score in the manuscript is this one.
\newcommand{\ThresholdCatalystPercent}{60}  % ThresholdCatalyst as a percentage, for prose.
\newcommand{\ThresholdTolerance}{0.20}  % The numeric tolerance: a number's penalty scales with relative error and saturates at this fraction.
\newcommand{\ThresholdTolerancePercent}{20}  % ThresholdTolerance as a percentage, for prose.
\newcommand{\ReleaseRecords}{3,942}  % row count
\newcommand{\ReleasePapers}{561}  % distinct doi
\newcommand{\ReleaseExcludedSolid}{1,120}  % source records whose catalyst is classed heterogeneous
\newcommand{\ReleaseExcludedUnknown}{501}  % source records whose catalyst resolves to no compound
\newcommand{\ReleaseAccepted}{2,661}  % audit == accepted
\newcommand{\ReleasePassRate}{67.5}  % accepted over judged, not over all rows
\newcommand{\ReleaseCorrected}{1,202}  % flagged and repairable: rejected with a proposed correction
\newcommand{\ReleaseCorrectedShare}{30.5}  % corrected as a share of judged
\newcommand{\ReleaseDropped}{79}  % judge set drop_record: rejected outright rather than repaired
\newcommand{\ReleaseDroppedShare}{2.0}  % dropped as a share of judged
\begin{document}

\pagestyle{fancy}
\thispagestyle{plain}
\fancypagestyle{plain}{
\renewcommand{\headrulewidth}{0pt}
}
\makeFNbottom
\makeatletter
\renewcommand\LARGE{\@setfontsize\LARGE{15pt}{17}}
\renewcommand\Large{\@setfontsize\Large{12pt}{14}}
\renewcommand\large{\@setfontsize\large{10pt}{12}}
\renewcommand\footnotesize{\@setfontsize\footnotesize{7pt}{10}}
\makeatother

\renewcommand{\thefootnote}{\fnsymbol{footnote}}
\renewcommand\footnoterule{\vspace*{1pt}% 
\color{cream}\hrule width 3.5in height 0.4pt \color{black}\vspace*{5pt}} 
\setcounter{secnumdepth}{5}

\makeatletter 
\renewcommand\@biblabel[1]{#1}            
\renewcommand\@makefntext[1]% 
{\noindent\makebox[0pt][r]{\@thefnmark\,}#1}
\makeatother 
\renewcommand{\figurename}{\small{Fig.}~}
\sectionfont{\sffamily\Large}
\subsectionfont{\normalsize}
\subsubsectionfont{\bf}
\setstretch{1.125}
\setlength{\skip\footins}{0.8cm}
\setlength{\footnotesep}{0.25cm}
\setlength{\jot}{10pt}
\titlespacing*{\section}{0pt}{4pt}{4pt}
\titlespacing*{\subsection}{0pt}{15pt}{1pt}
\fancyfoot{}
\fancyfoot[LO,RE]{\vspace{-7.1pt}\includegraphics[height=9pt]{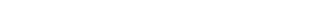}}
\fancyfoot[CO]{\vspace{-7.1pt}\hspace{13.2cm}\includegraphics{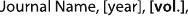}}
\fancyfoot[CE]{\vspace{-7.2pt}\hspace{-14.2cm}\includegraphics{head_foot/RF}}
\fancyfoot[RO]{\footnotesize{\sffamily{1--\pageref{LastPage} ~\textbar  \hspace{2pt}\thepage}}}
\fancyfoot[LE]{\footnotesize{\sffamily{\thepage~\textbar\hspace{3.45cm} 1--\pageref{LastPage}}}}
\fancyhead{}
\renewcommand{\headrulewidth}{0pt} 
\renewcommand{\footrulewidth}{0pt}
\setlength{\arrayrulewidth}{1pt}
\setlength{\columnsep}{6.5mm}
\setlength\bibsep{1pt}
\makeatletter 
\newlength{\figrulesep} 
\setlength{\figrulesep}{0.5\textfloatsep}

\newcommand{\topfigrule}{\vspace*{-1pt}% 
\noindent{\color{cream}\rule[-\figrulesep]{\columnwidth}{1.5pt}} }

\newcommand{\botfigrule}{\vspace*{-2pt}% 
\noindent{\color{cream}\rule[\figrulesep]{\columnwidth}{1.5pt}} }

\newcommand{\dblfigrule}{\vspace*{-1pt}% 
\noindent{\color{cream}\rule[-\figrulesep]{\textwidth}{1.5pt}} }

\makeatother
\twocolumn[
  \begin{@twocolumnfalse}
{\includegraphics[height=30pt]{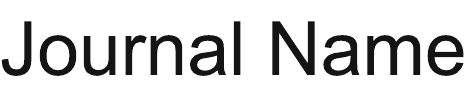}\hfill\raisebox{0pt}[0pt][0pt]{\includegraphics[height=55pt]{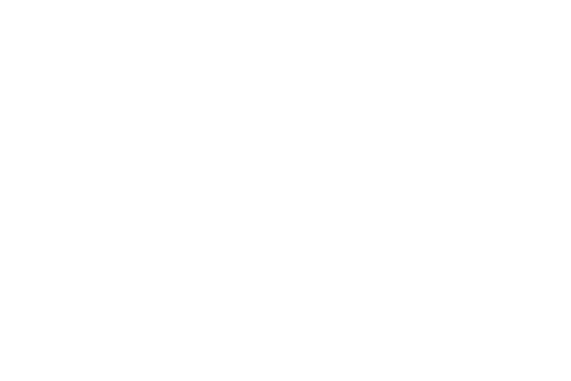}}\\[1ex]
\includegraphics[width=18.5cm]{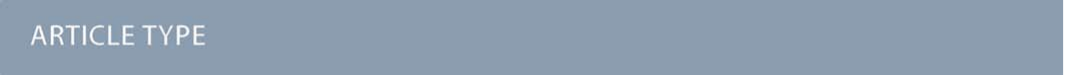}}\par
\vspace{1em}
\sffamily
\begin{tabular}{m{4.5cm} p{13.5cm} }

\includegraphics{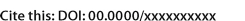} & \noindent\LARGE{\textbf{Building and Auditing Chemical Datasets with LLMs: A General LLM-as-a-Judge Framework Applied to Homogeneous Catalytic PET Depolymerisation$^\dag$}} \\
\vspace{0.3cm} & \vspace{0.3cm} \\

& \noindent\large{Muhammad bin Javaid,$^{\ast\ddag}$\textit{$^{a}$} Karim K. Ben Hicham,$^{\ddag}$\textit{$^{b}$} Enkhnyam Battulga,\textit{$^{a}$} Alexander Mitsos,\textit{$^{b,c,d}$} and Martin Grohe\textit{$^{a}$}} \\
\includegraphics{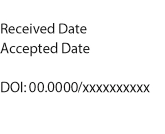} & \noindent\normalsize{Large language models can rapidly extract structured multivariate experimental data from the chemistry literature, yet auditing and verifying thousands of generated records without pre-existing ground truth remains difficult. Here, we present a general, citation-grounded extraction and verification methodology centered on an independent ``LLM-as-a-judge'', applied to the literature on homogeneous catalytic poly(ethylene terephthalate) (PET) depolymerisation. In this pipeline, an extractor LLM takes research papers as input and generates schema-enforced records with citations to document chunks tagged with universally unique identifiers (UUIDs); an independent LLM judge then reads the full paper alongside each record, flags unsupported fields, and proposes cited corrections. Prompt ablations demonstrate that requiring explicit chunk-level citations not only provides granular provenance down to specific paragraphs and tables, but also improves extraction accuracy.
To validate the judge before deploying it at scale, we benchmarked it against manual evaluation on records extracted from a hand-curated reference set of \CuratedExperiments{} ionic-liquid-catalysed PET glycolysis experiments. The judge agreed with human evaluation on \AdjJudgeAgrees{}\% of reviewed records and proved significantly more effective than a heuristic string-matching baseline at catching errors in extractions across related experiments. Deployed across the full corpus of \ReleaseRecords{} extracted records from \ReleasePapers{} papers, the framework audited and accepted \ReleasePassRate{}\% of them as written. Its proposed corrections concentrated on quantitative fields, where the extractor often conflated related experiments. We release the extracted dataset alongside the complete LLM judge audit logs. In addition, we release the \textit{Chemistry Data Extractor Toolkit}, an open-source application that implements this end-to-end pipeline, providing a code-free, browser-based environment for document parsing, customizable schema and prompt design, model-agnostic extraction and auditing, interactive provenance-highlighted review, and reproducible dataset bundling.}
\\

\end{tabular}

 \end{@twocolumnfalse} \vspace{0.6cm}
  ]

\renewcommand*\rmdefault{bch}\normalfont\upshape
\rmfamily
\section*{}
\vspace{-1cm}
\footnotetext{\textit{$^{a}$~RWTH Aachen University, Chair of Logic and Theory of Discrete Systems, Aachen, Germany. E-mail: \href{mailto:javaid@informatik.rwth-aachen.de}{javaid@informatik.rwth-aachen.de}}}
\footnotetext{\textit{$^{b}$~RWTH Aachen University, Process Systems Engineering (AVT.SVT), Aachen, Germany.}}
\footnotetext{\textit{$^{c}$~JARA-CSD, Aachen, Germany.}}
\footnotetext{\textit{$^{d}$~Forschungszentrum Jülich GmbH, Institute of Climate and Energy
Systems ICE-1: Energy Systems Engineering, Jülich, Germany}}
\footnotetext{\ddag~These authors contributed equally to this work.}
\footnotetext{\dag~Electronic supplementary information (ESI) available. See DOI: 00.0000/00000000.}
\section{Introduction}
Poly(ethylene terephthalate) (PET) is a widely used polyester in packaging and textiles, generating a substantial post-consumer waste stream for which effective recycling strategies are required. Chemical recycling by depolymerisation offers a route to recover monomers or other useful chemical feedstocks and can complement mechanical recycling, particularly for lower-quality or more heterogeneous PET waste streams \cite{mcneeleyAssessmentPETDepolymerization2024,bohreChemicalRecyclingProcesses2023}. Catalysis is central to these processes, and a wide range of homogeneous and heterogeneous systems have been developed for PET glycolysis, methanolysis, hydrolysis, and related transformations. Their reported performance, however, depends not only on catalyst identity but also on catalyst loading, depolymerising reagent or solvent, temperature, reaction time, PET feedstock, and product-isolation procedure, making comparisons between studies difficult \cite{bushUnderstandingImportantVariables2023}. Consolidating these experiments into structured datasets would enable systematic comparison of catalysts and reaction conditions, reveal gaps in the explored reaction space, and provide a foundation for statistical analysis and data-driven catalyst and process optimisation \cite{kearnesOpenReactionDatabase2021,yuOptimizingDiscoveryAIDriven2026}. In practice, the required information remains distributed across the prose, tables, figures, and supporting information of individual articles, such that manual extraction and standardisation are laborious and difficult to scale. Earlier chemistry-aware text-mining systems, including ChemDataExtractor and ChemDataExtractor 2.0, demonstrated the potential of automated extraction of chemical information and nested material--property relationships from scientific literature \cite{swain2016chemdataextractor,mavracic2021chemdataextractor}. Large language models (LLMs) offer a promising means of improving on this by being able to convert unstructured scientific literature into structured, schema-defined records, with recent studies demonstrating their ability to recover synthesis conditions, material properties, and chemical reaction information with only little or no task-specific training \cite{zhengChatGPTChemistryAssistant2023,polakExtractingAccurateMaterials2024,dagdelen2024structured,ai2024extracting,schilling-wilhelmiTextInsightLarge2025a}.

Despite this progress, improving the reliability of the extracted data remains a major challenge. Redundant prompting, chain-of-verification, and multi-agent reasoning generally operate within the extraction process and have primarily been evaluated on material--property tuples or synthesis descriptions \cite{polakExtractingAccurateMaterials2024,ansariAgentbasedLearningMaterials2024,liSLMMATRIXMultiagentTrajectory2025}. Schema-level judges can additionally identify malformed or internally inconsistent outputs \cite{ghoshAutomatedExtractionMaterial2026}, while recent LLM-as-a-judge pipelines evaluate extracted records against source documents using rubric-based assessments of dimensions such as completeness, condition extraction, and semantic accuracy \cite{zheng2023judging,lederbauerLeMatSynthMultimodalToolbox2025a}. However, such record- or criterion-level evaluations do not by themselves verify that each individual field is explicitly supported by the source and correctly associated with the same experimental instance, nor do they systematically provide source-linked, field-level corrections.

A related limitation concerns provenance. Work on citation-grounded generation has required LLM outputs to cite the passages supporting their claims \cite{gao2023enabling}. However, in scientific information-extraction workflows, the connection between a structured record and its precise supporting context is not always retained in a machine-addressable form~\cite{polakExtractingAccurateMaterials2024,ansariAgentbasedLearningMaterials2024}. Consequently, verifying an extracted value may require a human curator to search the article again, limiting the practical auditability of databases containing thousands of records. Retaining explicit references to the relevant text and table segments can instead make each record traceable throughout extraction, automated verification, and subsequent human review. Requiring the extractor to identify its supporting evidence may also act as a grounding constraint, although its effect on extraction accuracy has received comparatively little attention.

These requirements are particularly important for catalytic reaction literature. A single article may report an unordered series of closely related experiments that share catalyst and solvent identities while differing only in catalyst loading, reagent quantities, temperature, reaction time, or outcome. An extracted record can therefore be syntactically valid and chemically plausible while combining values from different experimental runs. Such association errors are difficult to detect using exact or fuzzy matching, especially when no comprehensive record-level reference dataset exists and no single field uniquely identifies an experiment. This also makes the human evaluation especially challenging.

Here, we develop and validate a source-cited and schema-enforced LLM data extraction framework and apply it to the literature on homogeneous catalytic PET depolymerisation. During preprocessing, every text and table chunk is assigned a unique identifier, and the extractor is instructed to list the identifiers of the chunks supporting each record. An independent LLM judge subsequently reads the source paper alongside the extracted record, evaluates whether its individual fields are supported and mutually associated with the same experimental run, and proposes source-chunk-linked corrections where necessary. We benchmark the judge against human assessment on a manually curated ionic-liquid subset before deploying it across a larger homogeneous-catalysis paper database. We further use prompt ablations to determine whether requiring explicit source references improves extraction accuracy in addition to enabling downstream traceability.

\section{Methods}

\begin{figure*}[h]
\centering
\includegraphics[width=\textwidth]{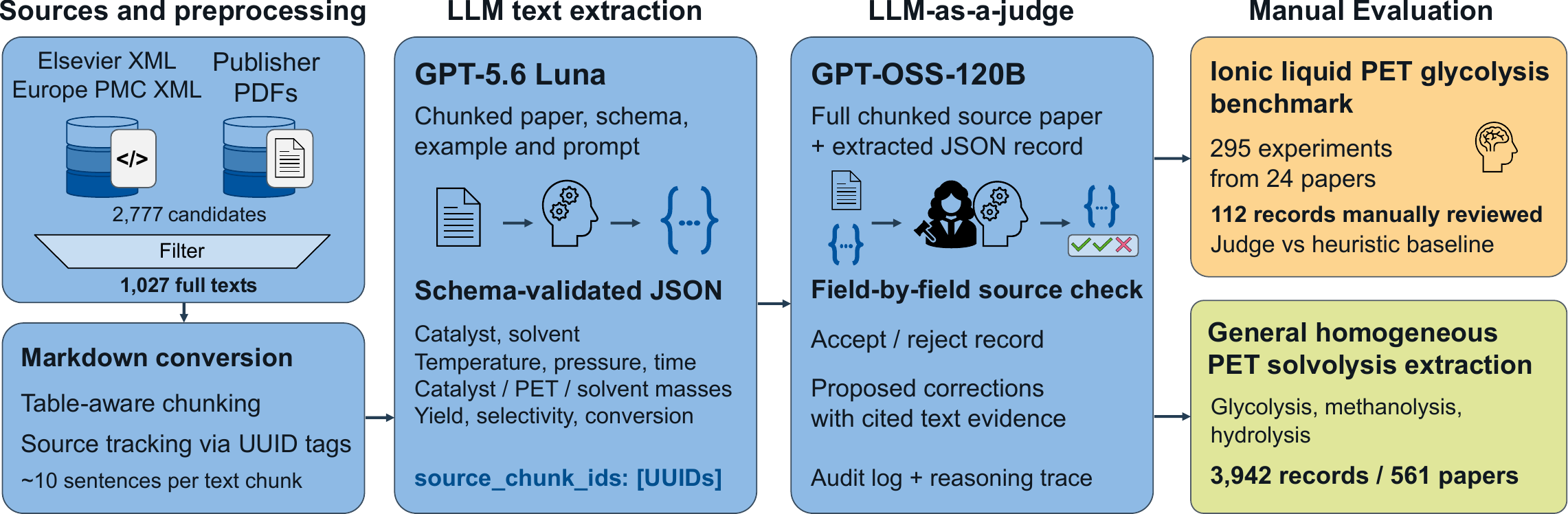}
\caption{Overview of the large language model (LLM) extraction and auditing pipeline for homogeneous catalytic poly(ethylene terephthalate) (PET) depolymerisation. Articles identified through Scopus are retrieved as XML or PDF files, converted to Markdown and segmented into text and table chunks with universally unique identifiers (UUIDs), preserving table structure. The extraction model (GPT-5.6 Luna) receives the chunked and UUID-tagged paper, a predefined schema, domain-specific guidelines, and one labelled worked example from an open-licensed glycolysis paper, excluding the target article. Schema-constrained extraction produces JSON records containing catalyst and solvent identities, reaction conditions, reactant amounts, and reaction outcomes. Each record includes a source chunk UUID field linking the extracted information to supporting source chunks. An independent LLM judge (GPT-OSS-120B) evaluates each record against the full chunked paper, issues an accept/reject verdict and proposes corrections supported by cited text evidence and a reasoning trace. Judge validation uses an ionic liquid benchmark comprising \CuratedExperiments{} curated experiments from \CuratedPapers{} papers, with manual assessment of \AdjRecords{} extracted records and comparison against a heuristic grader. The resulting dataset covers PET glycolysis, methanolysis and hydrolysis, retaining raw extractions alongside audit logs and proposed corrections.}
\label{fig:overview}
\end{figure*}

\subsection{Extraction Schema}
We target three depolymerisation routes for PET that share a consistent reaction format: glycolysis, methanolysis, and hydrolysis. In each route, PET reacts with a solvent in the presence of a catalyst at a specified temperature and pressure, for a given duration.

The extraction schema defines 12 fields per experimental record: catalyst, solvent, temperature, reaction time, three distinct mass values (catalyst, PET, solvent), yield, selectivity, conversion, pressure, and the source chunk identifiers \cite{gao2025mlpet, kumar2024energy}. The reaction route itself is not extracted but is deterministically inferred from the extracted solvent name (e.g., ethylene glycol and related diols designate glycolysis, methanol designates methanolysis, and water or aqueous acids/bases designate hydrolysis). Consequently, the reaction product field is deliberately omitted because it is fixed by this inferred route (e.g., bis(2-hydroxyethyl) terephthalate for glycolysis, dimethyl terephthalate for methanolysis, and terephthalic acid for hydrolysis).

\subsection{Document Retrieval and Preprocessing}
Candidate articles were identified via Scopus literature searches, and full-text documents were retrieved across multiple sources: Elsevier XML articles via Elsevier's Text and Data Mining API, open-access articles from Europe PMC in JATS XML format, and PDFs from other publishers (primarily Wiley and Springer) retrieved programmatically via their TDM services, institutional entitlements, and the Unpaywall API for open-access content.

To standardise these formats into a uniform Markdown representation, we utilised a custom parser \cite{splittermr_fork} built upon SplitterMR \cite{splittermr} for Elsevier XML, a JATS reader for Europe PMC articles, and Docling \cite{docling2024, docling} for PDFs.

To enable precise source tracking, the preprocessing pipeline applies table-aware chunking. Each document is segmented into semantic blocks, with standard text divided into roughly ten-sentence, non-overlapping chunks, while retaining the preserved row-column structures of extracted tables. Visual elements, such as graphs and figures, are excluded due to the high error rates of current optical character recognition (OCR) models when interpreting complex chemical notations. Every generated text or table chunk is assigned a universally unique identifier (UUID). This UUID referencing system ensures that any data point extracted by the LLM can be deterministically traced back to its exact location in the source document.

\subsection{The LLM Extraction and Verification Pipeline}
The extraction workflow is separated into two distinct phases. A full overview is given in Fig.~\ref{fig:overview}.

\textbf{Phase 1: Primary Extraction.} The generator LLM is provided with the chunked document, the schema definition, domain-specific extraction guidelines, and a single labelled worked example. For the final data extraction campaign, we deployed \texttt{gpt-5.6-luna} (accessed via Azure OpenAI) as the primary extractor. To guarantee that the output is syntactically valid and strictly adheres to the required data types, we utilise API-level structured output formatting \cite{openai2024structured} managed via LiteLLM \cite{litellm}. The LLM is directly constrained to generate data conforming to a predefined JSON schema, which the API runtime validates before returning the response.
The extraction prompt defines scope boundaries and domain rules: it instructs the model to capture all primary catalytic runs from text and tables while explicitly skipping cited literature benchmarks, response-surface/optimization grids, and characterization tables. It further directs the model to standardize catalyst formulas into bracket notation, propagate global conditions stated in methods or table footnotes to all corresponding runs, compute absolute reagent masses from reported weight percentages or molar ratios, and strictly distinguish between yield, conversion, and selectivity metrics (the full prompt is provided in Section~S2$\dag$, ESI).

The prompt enforces explicit source tracking by requiring the model to populate a \texttt{source\_chunk\_ids} field with the exact UUIDs of the referenced text chunks. Beyond ensuring data provenance, this instruction might act as an implicit attention anchor. Ablations demonstrated that requiring the model to point to specific chunks increases the extraction's $F_1$ score from $\sim$\SourceNot{} to $\sim$\SourceCiting{}. We include exactly one worked example; sweeps demonstrated that this performs just as well as four or five examples, increasing extraction quality while saving input tokens per paper. Exemplars are drawn from a pool of open-licensed benchmark papers, strictly excluding the target paper to prevent data leakage. Since LLM generation is inherently stochastic, benchmark figures and ablations were averaged over three independent repeats to account for run-to-run variance.

\textbf{Phase 2: LLM-as-a-Judge.} An independent LLM acts as an automated auditor. For our production pipeline, we deployed \texttt{gpt-oss-120b}. Hosted on an institutional endpoint, this model provided an unmetered \JudgeContextCap{}-token context window, making exhaustive auditing computationally affordable without token limits truncating long documents. It is supplied with both the full chunked paper and the structured JSON record generated in Phase 1. The judge model evaluates the record field-by-field against the source text, writes a step-by-step reasoning trace, issues a final verdict (accept or reject), and proposes specific corrected values alongside cited text evidence for any field it deems unsupported by the source text.

The judge prompt instructs the LLM to evaluate chemical identity, accepting standard abbreviations and IUPAC synonyms while rejecting non-standard, opaque abbreviations defined only in the paper itself. It also instructs to ensure the required condition propagation and unit conversions were accurately performed, to check mathematical consistency across reported outcomes (e.g., $\text{yield} = \text{conversion} \times \text{selectivity} / 100$), and to disregard harmless formatting variations. To prevent ungrounded verdicts, the model is required to write a 2--5 sentence evidence critique citing chunk UUIDs before outputting a verdict and proposed field changes (the full judge prompt is provided in Section~S2$\dag$, ESI).

\subsection{Validation and Baseline Comparison}

To audit extracted records from thousands of papers where no ground truth in the form of a curated dataset exists, an automated grader must be able to evaluate the faithfulness of the records with respect to the source texts directly. Before applying the LLM-as-a-judge framework to the full \ReleaseRecords{}-record database, we evaluated it against a string-matching heuristic on a hand-curated reference set. While the heuristic evaluates extractions against human-curated reference tables, the LLM judge evaluates extractions directly against the raw paper text.

\textbf{Curated Benchmark Dataset.} This reference set comprises \CuratedExperiments{} experiments from \CuratedPapers{} papers on ionic liquid-catalysed PET glycolysis, drawing from datasets previously assembled by Gao \textit{et al.}\cite{gao2025mlpet} and Kumar \textit{et al.}\cite{kumar2024energy}. Because our document preprocessing pipeline excludes graphical figures and charts, experimental runs reported exclusively in graphical form were omitted from this reference set, thus avoiding penalising extractions for missing visual-only data.

\begin{figure*}[h]
\centering
\includegraphics[width=\textwidth]{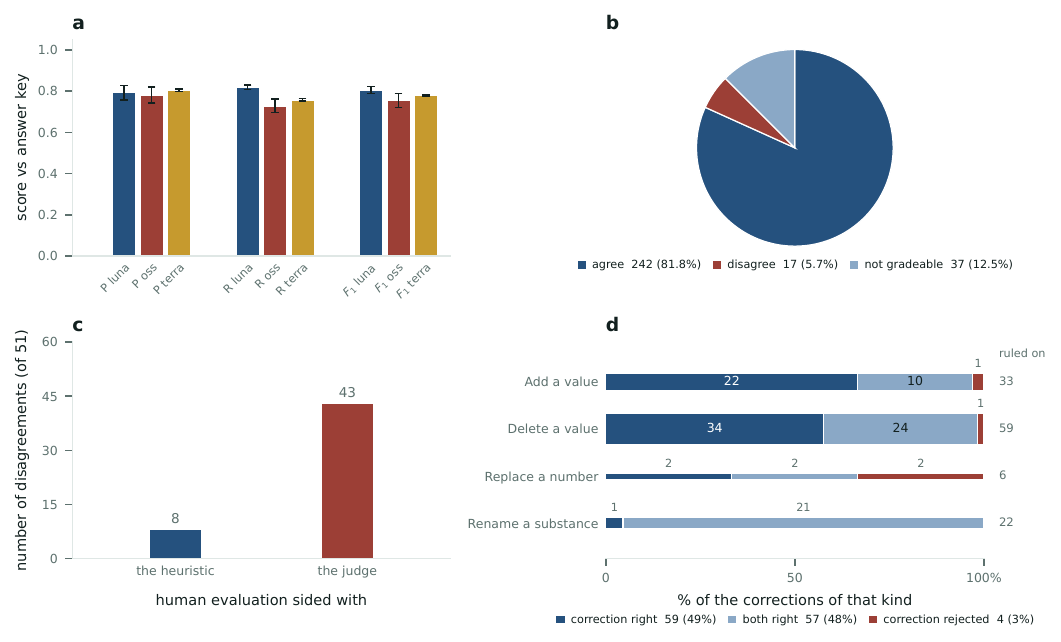}
\caption{\textbf{Extraction performance and automated grader validation.} (\textbf{a}) Precision ($P$), recall ($R$), and $F_1$ scores across candidate extraction models evaluated against the curated benchmark dataset. (\textbf{b}) Grader consensus across \MatrixRecords{} benchmark records. The 81.8\% slice represents mutual agreement on evaluable records (242/259, 93.4\%); total agreement reaches 82.8\% when accounting for three uncurated records mutually rejected by both graders.
(\textbf{c}) Manual evaluation outcomes across the \AdjPairs{} informative disagreement pairs, where \AdjFavourJudge{} cases favored the judge and only \AdjFavourMetric{} favored the heuristic. (\textbf{d}) Human evaluation of judge-proposed corrections by edit type across 120 reviewed interventions, showing that 97\% were either demonstrably correct (49\%) or represented equally valid reporting variants (``both right'', 48\%), with only 4 proposals (3\%) rejected as erroneous.}
\label{fig:graders}
\end{figure*}

\textbf{Model Selection.} To determine the optimal configuration for mass data extraction, we benchmarked three models: \texttt{gpt-5.6-luna}, \texttt{gpt-oss-120b}, and \texttt{gpt-5.6-terra}, evaluating all nine possible extractor-judge combinations against the curated reference set (Table~S1$\dag$, ESI). The deployed pairing of \texttt{luna} for extraction and \texttt{oss} for judging was deliberately chosen for achieving a high $F_1$ score alongside an affordable, unmetered judging step (Fig.~\ref{fig:graders}a and Fig.~S3f$\dag$, ESI).

\textbf{The Heuristic Grader.} Most materials extraction benchmarks pair predictions to ground truth using a single identifier, such as a polymer name or chemical formula \cite{polakExtractingAccurateMaterials2024,guptaDataExtractionPolymer2024a}. However, in our case, because automated extraction yields an unordered set of $m$ candidate records from a paper with $n$ curated reference experiments, evaluation cannot rely on order of appearance or on a single primary key (such as catalyst identity), as individual papers frequently report multiple runs sharing identical catalysts and solvents. To our knowledge, no standard benchmark heuristic exists in the chemical literature for evaluating unordered, multi-run reaction records. We therefore formulated record evaluation as an optimal bipartite matching problem solved via the Hungarian algorithm \cite{kuhn1955hungarian}, combining combinatorial assignment with a custom continuous penalty metric designed to avoid the brittleness of binary exact matching.

The heuristic evaluates 10 of the 12 schema fields: catalyst name, solvent name, temperature, reaction time, catalyst mass, PET mass, solvent mass, yield, selectivity, and conversion. Pressure is excluded from scoring because the vast majority of curated experiments were conducted at ambient pressure (1\,atm), offering little meaningful variation across runs, while source chunk identifiers are excluded as internal provenance metadata. For each field $k$, a penalty $p_k \in [0, 1]$ is computed between candidate value $x_{\text{ext}}$ and reference value $x_{\text{cur}}$:
\begin{equation}
    p_k = \begin{cases}
        0, & x_{\text{ext}} = x_{\text{cur}} = \text{null}, \\
        1, & \text{either } x_{\text{ext}} \text{ or } x_{\text{cur}} = \text{null}, \\
        \min\left(\frac{|x_{\text{ext}} - x_{\text{cur}}|}{\ThresholdTolerance\, x_{\text{cur}}}, 1\right), & \text{numeric fields}, \\
        \mathbb{I}(\text{fuzzy string similarity} < \ThresholdCatalyst ), & \text{catalyst}, \\
        \mathbb{I}(x_{\text{ext}} \neq x_{\text{cur}}), & \text{solvent},
    \end{cases}
\end{equation}
where $\mathbb{I}(\cdot)$ is the indicator function. This penalty structure reflects the differing reporting conventions and error tolerances of each field. When both the extraction and the reference agree a parameter was unrecorded (null), the penalty is zero, whereas a missed or hallucinated value incurs the maximum penalty of 1. For numeric values, scoring by relative error up to a \ThresholdTolerancePercent{}\% cap tolerates routine rounding and differences in significant figures while treating substantial discrepancies as outright errors. For catalysts, fuzzy string similarity (computed after lowercasing, whitespace stripping, and bracket standardisation) accommodates the varied notation common across homogeneous catalyst nomenclature, such as inconsistent spacing, ligand abbreviations, or bracket variations like $[\text{bmim}]\text{Cl}$ versus $[\text{Bmim}][\text{Cl}]$. Conversely, solvents determine the depolymerisation route itself and comprise standard reagent names, warranting an exact string match. The overall penalty for a candidate pair is the unweighted mean across all 10 evaluated fields: $\bar{p} = \frac{1}{10} \sum_{k=1}^{10} p_k$.

The Hungarian algorithm \cite{kuhn1955hungarian} is applied to the resulting $m \times n$ cost matrix to find the one-to-one assignment that globally minimizes total penalty across the paper's candidate and reference records. An assigned pair qualifies as a true positive if and only if two conditions are met: the catalyst names share at least \ThresholdCatalystPercent{}\% fuzzy string similarity, and the mean record penalty satisfies $\bar{p} < \tau$ ($\tau = \ThresholdAccept{}$). Sensitivity analysis confirmed that $\tau = \ThresholdAccept{}$ sits on a stable plateau; because faithful extractions typically exhibit average penalties well below \ThresholdAccept{}, raising the threshold further has negligible impact on the final $F_1$ score.

Pairs that fail either condition yield both a false negative (the ground-truth experiment was missed) and a false positive (the extracted run was invalid). Likewise, any extractions left without a partner count as false positives, and any reference experiments the extractor overlooked count as false negatives.

\textbf{Manual Evaluation Protocol.} A sample of \AdjRecords{} extracted records from the ionic liquid subset was manually evaluated to determine if they were faithful representations of the source papers. To maximize statistical power, the human evaluation followed a stratified sampling design: records where the two automatic graders disagreed were exhaustively evaluated, while a subset of records they agreed upon was randomly sampled. Both the LLM judge and the heuristic grader were scored against these human verdicts. McNemar's test \cite{mcnemar} was then applied to the conflicting pairs, i.e., instances where only one grader agreed with human evaluation, to establish the statistical reliability of the LLM judge prior to its deployment on the broader database.

\begin{figure*}[h]
\centering
\includegraphics[width=\textwidth]{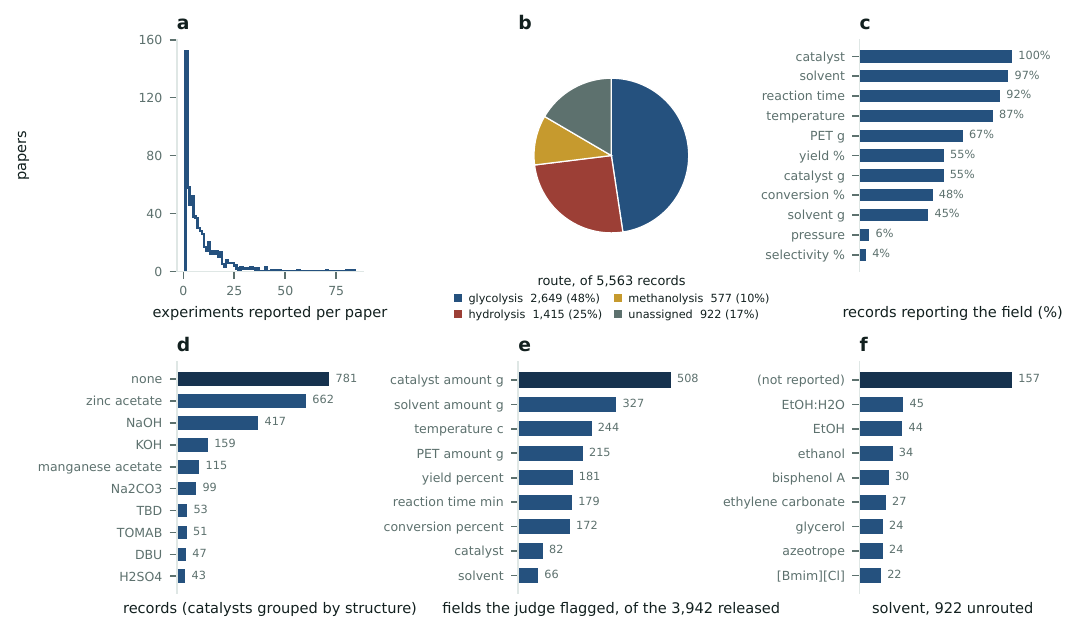}
\caption{\textbf{Dataset composition, field completeness, and LLM judge interventions.} (\textbf{a}) Distribution of extracted experiments per paper. (\textbf{b}) Breakdown across depolymerisation routes, led by glycolysis (\RouteGlycolysisShare{}\%) and hydrolysis (\RouteHydrolysisShare{}\%). (\textbf{c}) Completeness by schema field. (\textbf{d}) Most frequent catalyst entries, dominated by uncatalysed baselines, zinc acetate, and simple alkaline catalysts. (\textbf{e}) Field-level correction counts for nine of the eleven fields (excl. selectivity, pressure) proposed by the LLM judge, concentrated heavily on catalyst and solvent amounts (restricted to the \ReleaseRecords{} homogeneous catalysis subset). (\textbf{f}) Solvent identities for the \RouteOther{} unrouted records from (\textbf{b}), accounting for unconventional reaction media and omitted solvent names.}
\label{fig:database}
\end{figure*}

\begin{figure*}[htbp]
\centering
\includegraphics[width=\textwidth]{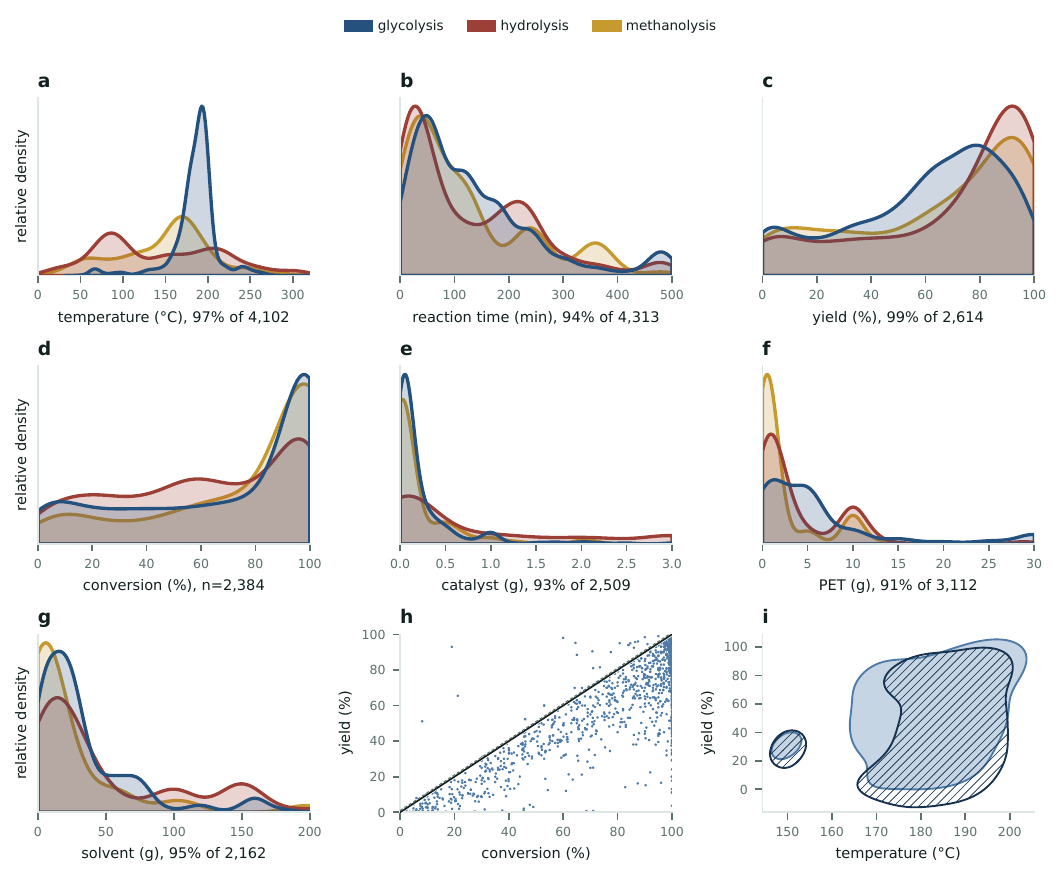}
\caption{\textbf{Reported reaction conditions statistics and chemical consistency checks.} 
(\textbf{a}--\textbf{g}) Kernel density estimation (KDE) plots of key experimental parameters broken down by depolymerisation route (blue: glycolysis; red: hydrolysis; yellow: methanolysis). Panels (a--g) cover the \RouteNamedTotal{} route-assigned records. Each label gives how many report the field, preceded where relevant by the percentage falling inside the plotted range; this is not field completeness, which Fig.~3c provides.
(\textbf{a}) Reaction temperature, exhibiting a sharp glycolysis peak at 190--200\,\textdegree C corresponding to the atmospheric reflux of ethylene glycol. 
(\textbf{b}) Reaction time, clustered at standard bench protocol intervals.
(\textbf{c}) Product yield and (\textbf{d}) PET conversion, both skewed toward 100\%. 
(\textbf{e}) Catalyst mass, (\textbf{f}) PET feedstock mass, and (\textbf{g}) solvent mass, concentrated at standard laboratory scales. 
(\textbf{h}) Physical consistency check between yield and conversion against the 1:1 parity line (dashed; $\text{yield} \le \text{conversion}$), with only \IdentityImpossible{} of \IdentityPairs{} records violating the physical bound. 
(\textbf{i}) Bivariate density contour comparison of temperature versus yield between this work (solid blue) and the ionic liquid depolymerisation dataset by Gao \textit{et al.}\cite{gao2025mlpet} (hatched) for the \GaoPapers{} articles common to both datasets (extended field comparisons in Fig.~S1$\dag$, ESI).
}
\label{fig:chemistry}
\end{figure*}

\section{Results and discussion}

\subsection{Model Selection and LLM Judge Verification}

To inform the choice of model for data extraction, we evaluated three LLMs in terms of their performance on \CuratedPapers{} benchmark papers (\CuratedExperiments{} hand-curated experiments): GPT-5.6 Luna, GPT-OSS-120B, and GPT-5.6 Terra (Fig.~\ref{fig:graders}a). GPT-5.6 Luna demonstrated the best mean recall and $F_1$ score while achieving precision comparable to the other two models, so it was chosen for the final mass data extraction.

For the judge model, we employed the open-weights model GPT-OSS-120B. While our heuristic grader provides an automated check on benchmark data, scaling it to thousands of uncurated papers is impossible without pre-existing reference keys. To evaluate whether the LLM judge could perform this auditing role without a reference key, we compared its judgments against the heuristic grader on the benchmark papers. Restricted to evaluable records, i.e., those with a direct counterpart in the curated benchmark, the judge and heuristic exhibit high agreement at 93.4\% (242/259) for the deployed model pairing (GPT-5.6 Luna extractor and GPT-OSS-120B judge). Across all 296 records, overall agreement is 82.8\% (245/296), which includes three extraneous extractions that both graders independently rejected. Fig.~\ref{fig:graders}b shows a breakdown of the agreement between LLM judge and heuristic grader. However, this metric alone is not enough: high agreement could also be obtained in the case where both graders simply make similar errors. This necessitates manual evaluation to assess the quality of the LLM judgments.

\subsection{Agreement with Manual Evaluation}

To benchmark the automated audit against human judgment, we conducted a manual evaluation of \AdjRecords{} sample extracted records against their source articles. We employed a stratified sampling strategy, reviewing all \AdjPairs{} records where the two automated graders disagreed and all \AdjBothFlagged{} records where both raised flags, alongside a random sample of \AdjAcceptedReviewed{} records that both accepted. Of these \AdjRecords{} records, human evaluation agreed with the LLM judge's verdict on 96 of them (\AdjJudgeAgrees\%). We note that all of the \AdjAcceptedReviewed{} sampled records that both the LLM judge and heuristic grader accepted were verified to be correct by manual evaluation.

Manual review revealed that the extracted records were largely faithful to the source literature, with apparent discrepancies frequently reflecting benign formatting and reporting variants rather than genuine extraction failures. 
When evaluating the judge's field-level proposals across 120 reviewed interventions (Fig. \ref{fig:graders}d), 59 (49\%) were confirmed as demonstrably correct fixes and 57 (48\%) represented equally valid reporting conventions (classified as ``both right''), such as chemical synonymy or standard parameter omissions. 
Only 4 proposed corrections (3\%) were rejected as outright erroneous; in summary, 97\% of the judge proposals were either a real fix or harmless.
In contrast, the heuristic grader flagged records far more indiscriminately, penalizing valid extractions due to rigid string-matching thresholds and an inability to recognize non-identical but chemically equivalent descriptions. 

In head-to-head comparison, the judge substantially outperformed the heuristic baseline (Fig.~\ref{fig:graders}c). Across the \AdjPairs{} discordant records where only one grader agreed with manual evaluation, \AdjFavourJudge{} cases favored the LLM judge and only \AdjFavourMetric{} favored the heuristic, a difference significant at $p = \AdjMcNemarP$ by McNemar's test. The LLM judge therefore provides a statistically grounded triage tool that effectively concentrates manual review on records most likely to require correction.

Because manual review confirmed that the mutual acceptances were all valid records and that disagreements decisively favored the judge, we concluded that the LLM judge matches or exceeds the reliability of a reference-based heuristic while reading the source literature directly. This validated deploying the judge as an automated auditor for the subsequent uncurated mass data extraction.

\subsection{Prompt Ablations}

Two prompt settings were swept across the \CuratedPapers{} curated benchmark papers (three independent repeats each).

First, we tested the number of worked examples included in the prompt. Providing a single labelled example raised the $F_1$ score by $\ShotsDelta{}$ compared to zero-shot extraction ($p = \ShotsP{}$). Adding further examples (up to five) yielded no significant improvements beyond noise, indicating that one example provides sufficient formatting guidance while saving a significant number of tokens per paper.

Second, requiring each record to cite its source chunks yielded an $F_1$ improvement of $\SourceDelta{}$ ($\SourceCiting{} \pm \SourceCitingSd{}$ with sourcing vs. $\SourceNot{} \pm \SourceNotSd{}$ without; $p = \SourceP{}$). Although both effects sit just shy of the standard $\alpha=0.05$ significance threshold due to the run-to-run variance across the three repeats, the consistent positive direction supports maintaining both one worked example and explicit source tracking in the production pipeline. Beyond this empirical benefit, chunk-level citations establish end-to-end data provenance, anchoring every extracted reaction parameter in the source text to enable quick manual verification and trustworthy downstream reuse.

\subsection{Building the Dataset at Scale}

From an initial pool of \CorpusCandidates{} candidate articles identified via Scopus, \CorpusFiltered{} relevant papers passed title and abstract screening (Table~\ref{tab:funnel}). The drop from \CorpusFiltered{} to \CorpusConverted{} was due to access issues. We obtained \CorpusElsevierXml{} Elsevier papers and a subset of Wiley/Springer papers via text-and-data-mining entitlements, alongside open-access PDFs and Europe PMC JATS deposits. The \CorpusNotObtained{} missing papers are heavily concentrated in ACS and RSC subscription content.

\begin{table}[h]
\small
\caption{From search to corpus.}
\label{tab:funnel}
\begin{tabular*}{\columnwidth}{@{\extracolsep{\fill}}lr}
\toprule
Selection stage & Articles \\
\midrule
Candidates identified via Scopus searches       & \CorpusCandidates{} \\
\quad Excluded: no polymer named              & $-\FunnelDroppedPolymer{}$ \\
\quad Excluded: no depolymerisation route named & $-\FunnelDroppedRoute{}$ \\
\quad Excluded: off-topic subjects or reviews   & $-\FunnelDroppedOffTopic{}$ \\
Articles passing title and abstract filter    & \CorpusFiltered{} \ (\FunnelPassRate{}\%) \\
\quad Obtained: Elsevier full-text XML        & \CorpusElsevierXml{} \\
\quad Obtained: PDF (Wiley, Springer, OA)     & \CorpusPdf{} \\
\quad Obtained: Europe PMC JATS               & \CorpusEuropePmc{} \\
\quad Obtained: other routes                  & \CorpusOtherSource{} \\   
Retrieved and converted to chunked text      & \textbf{\CorpusConverted{}} \\
\bottomrule
\end{tabular*}
\end{table}

Of the \CorpusConverted{} converted papers, \DatabasePapers{} were successfully processed by the extraction pipeline (one excessively long paper exhausted its rate-limit retries). Table~\ref{tab:massextraction} summarises the final mass extraction. Every output parsed into valid JSON, demonstrating the robustness of the schema and API-level tool-use constraints. While the initial extraction recovered \DatabaseRecords{} records across all reported catalytic runs, our study focuses strictly on homogeneous systems. Filtering out \ReleaseExcludedSolid{} records involving heterogeneous (solid) catalysts and \ReleaseExcludedUnknown{} with unverified catalyst phases yielded the final dataset of \ReleaseRecords{} homogeneous catalytic experiments across \ReleasePapers{} articles.

Most of the \DatabaseEmpty{} papers that yielded no records were simply out of scope. Title inspection classified \EmptyBiological{} as concerning biological degradation, \EmptyOffTarget{} as following off-target routes or using PET solely as a material, and \EmptyReview{} as reviews without protocols, leaving roughly \EmptyUnclassified{} unclassified. While we initially suspected context-window limits on long documents, re-running the 15 largest papers confirmed they were merely long-form reviews and protocols devoid of extractable experimental runs.

\begin{table}[h]
\small
\caption{Mass extraction and curation summary.}
\label{tab:massextraction}
\begin{tabular*}{\columnwidth}{@{\extracolsep{\fill}}lr}
\toprule
Metric & Value \\
\midrule
Articles processed        & \DatabasePapers{} \\
Articles yielding records & \DatabaseYielding{} \\
Articles yielding no records & \DatabaseEmpty{} \\
Unparseable outputs     & 0 \\
Extraction cost         & \$\DatabaseCost{} \\
\midrule
Total extracted experimental records & \DatabaseRecords{} \\
\quad Excluded: heterogeneous (solid) catalysts & $-\ReleaseExcludedSolid{}$ \\
\quad Excluded: unverified catalyst phase       & $-\ReleaseExcludedUnknown{}$ \\
Final homogeneous records released              & \textbf{\ReleaseRecords{}} \\
\bottomrule
\end{tabular*}
\end{table}

\subsection{Dataset Characteristics and Verification}

Most papers reported between 1 and 25 experiments, with glycolysis being the route in \RouteGlycolysisShare{}\% of the records extracted in the dataset, followed by hydrolysis (\RouteHydrolysisShare{}\%) and methanolysis (\RouteMethanolysisShare{}\%). The remainder were unassigned routes that did not fall under these three categories. Summary statistics for the dataset fields are shown in Fig.~\ref{fig:database}, and the resulting distributions of reaction conditions, feedstock amounts, and outcomes are shown in Fig.~\ref{fig:chemistry}a--g.

After data extraction and filtering to homogenous catalysis, the LLM judge read every source paper and evaluated all \ReleaseRecords{} generated records, accepting \ReleasePassRate\% exactly as written (Table~\ref{tab:verdicts}). The judge proposed corrections for \ReleaseCorrectedShare\% of the records.

\begin{table}[h]
\small
\caption{The judge's verdicts on the released dataset.}
\label{tab:verdicts}
\begin{tabular*}{\columnwidth}{@{\extracolsep{\fill}}lrr}
\toprule
Verdict category & Records & Share \\
\midrule
Accepted as written                & \ReleaseAccepted  & \ReleasePassRate\% \\
Rejected, with proposed correction & \ReleaseCorrected & \ReleaseCorrectedShare\% \\
Rejected outright                  & \ReleaseDropped   & \ReleaseDroppedShare\% \\
\midrule
Total                              & \ReleaseRecords   & 100.0\% \\
\bottomrule
\end{tabular*}
\end{table}

These proposed corrections were heavily concentrated in the quantitative mass fields (Fig.~\ref{fig:database}e). Catalyst mass (508 fixes) and solvent mass (327 fixes) dominate the error distribution. As authors typically state absolute masses once in a methods paragraph and then vary them implicitly as ratios across a results table, an extractor reading the table without properly carrying the paragraph context forward may capture the ratios correctly but miscalculate the absolute amounts.

We can further evaluate data quality against physical constraints (Fig.~\ref{fig:chemistry}h). By definition, reaction yield cannot exceed conversion; across the corpus, only \IdentityImpossible{} out of \IdentityPairs{} records reporting both metrics violate this boundary, confirming that the extractor rarely conflates the two quantities.

For researchers utilizing this database for predictive modeling, accounting for study-level fixed effects is essential. Because published experiments span diverse PET feedstock forms, catalyst activities, and experimental scales, pooling raw records globally across the corpus obscures fundamental chemical trends, yielding, for example, virtually no overall correlation between temperature and yield ($\rho = \WithinPooledRho{}$). However, evaluating each study individually resolves these cross-study confounds. Within individual papers, temperature and yield exhibit a consistent positive relationship, with a median intra-study correlation of $\rho = +\WithinPaperRho{}$ and \WithinPaperPositive{}\% of multi-run papers displaying positive trends ($p = \WithinPaperP{}$ via Wilcoxon signed-rank test). We note that the overall relationship between temperature and yield across papers in our dataset is similar to that in the manually curated dataset on ionic liquid-based PET waste recycling by Gao \textit{et al.}\cite{gao2025mlpet} (Fig.~\ref{fig:chemistry}i), a correspondence that holds across further key experimental parameter pairs (Fig.~S1$\dag$, ESI).

Finally, of the \CitationsTotal{} source chunk citations generated by the extractor, \CitationsResolved{} (\CitationsTraceable{}\%) resolved correctly to a valid chunk within the corresponding paper. Only a minority failed to match, primarily due to verbatim copying of the prompt's worked examples. In the interests of transparency and data provenance, we release the LLM judge's proposed corrections alongside the raw extractions and complete reasoning traces.

\begin{figure*}[htbp]
\centering
\includegraphics[width=\textwidth]{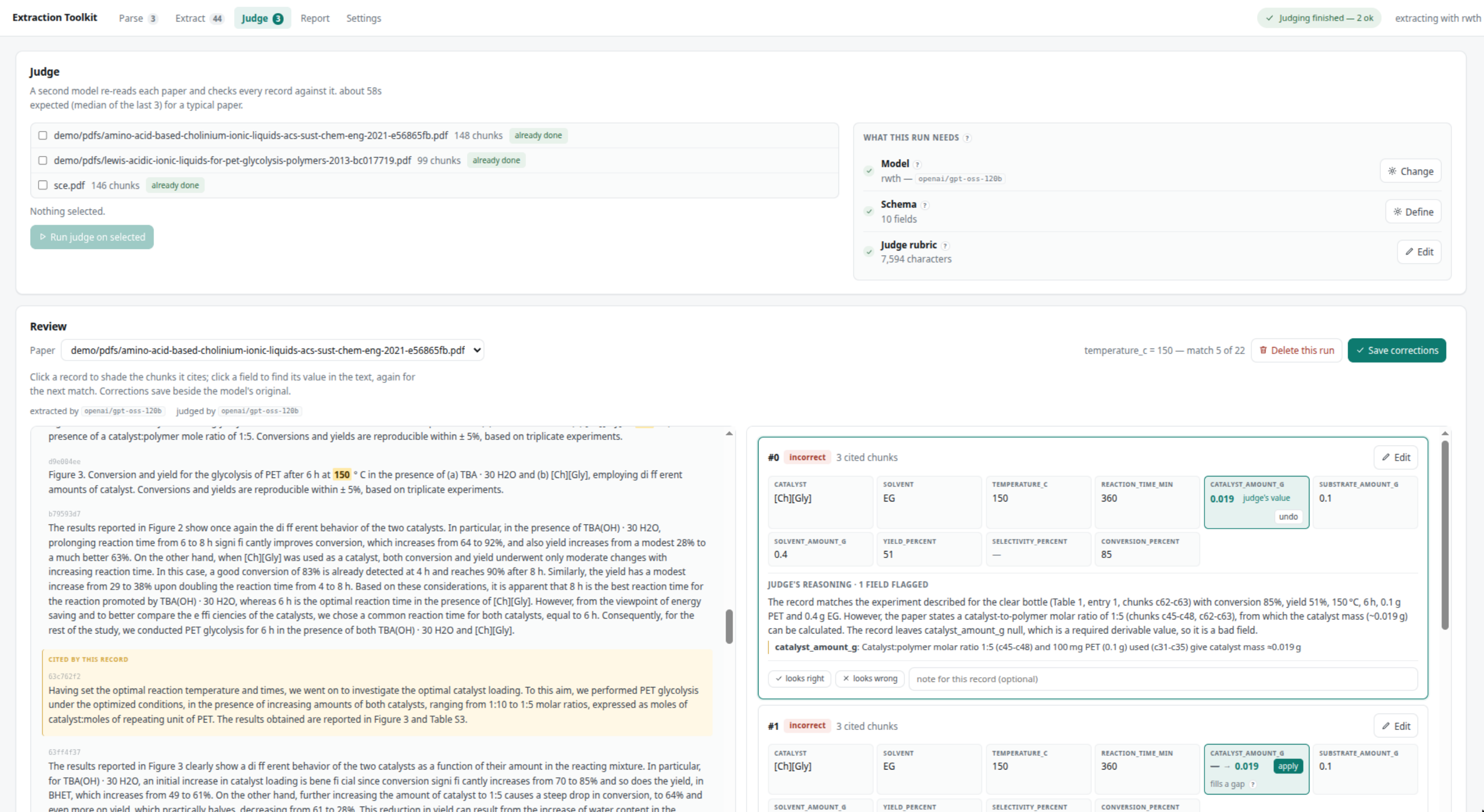}
\caption{\textbf{Screenshot of the interactive Chemistry Data Extractor Toolkit.} The local browser-based interface enables end-to-end dataset curation and auditing without code modification. Users configure target schemas, extraction prompts, and judge rubrics directly in the settings. The dual-pane review interface links extracted records to the source paper, highlighting cited UUID chunks upon selection and allowing one-click inspection of the judge's proposed corrections and reasoning.}
\label{fig:toolkit}
\end{figure*}

\subsection{Interactive Extraction and Auditing Toolkit}

To make the citation-grounded extraction and judging workflow accessible and reusable across other chemical domains, we implemented the framework as an open-source, local-first web application: the \textit{Chemistry Data Extractor Toolkit} (Fig.~\ref{fig:toolkit}). The software runs locally via Docker or Python (\texttt{uv}) without external database dependencies, storing project state, PDFs, and extracted records in local workspace directories.

The application is fully configurable through a browser interface, allowing users to adapt the pipeline to new catalytic reactions, materials, or properties without modifying the codebase. Users define target fields and descriptions in a schema editor, provide custom extraction guidelines and worked examples, and formulate judging rubrics for automated auditing. The pipeline is model-agnostic, supporting OpenAI, Anthropic, Azure, and local endpoints (via Ollama or vLLM), allowing users to pair high-throughput extraction models with independent open-weights or frontier judge models.

Integrated document parsing (powered by Docling) segments dropped PDFs into Markdown text and structured tables tagged with chunk UUIDs. During human-in-the-loop review (Fig.~\ref{fig:toolkit}), the interface presents a synchronized two-pane view: selecting an extracted record highlights its cited UUID text chunks in the source paper on the left, while clicking an individual field jumps directly to matching values in the source text. The interface integrates the judge's reasoning and flags, allowing human curators to accept, edit, or reject proposed corrections with a single click. Finally, to ensure reproducibility, the tool exports a complete dataset bundle containing the extracted records alongside the active schema, prompts, cited chunk mappings, per-paper model metadata, token costs, and the software commit hash.

\section{Limitations}

Our benchmark relies on \CuratedExperiments{} hand-curated experiments across \CuratedPapers{} papers, all covering PET glycolysis. Although the extraction schema applies equally well to methanolysis and hydrolysis, we did not benchmark those two routes against a hand-curated reference key. Because our prompt examples were drawn entirely from glycolysis papers, the model may perform slightly differently when dealing with other solvents and reaction setups.

The audit is also designed to check accuracy, not completeness. The LLM judge flags unsupported numbers in records the extractor produced, but it cannot catch experiments the extractor missed entirely. In addition, the judge raises some false alarms, meaning automatic corrections could introduce new mistakes. Therefore, we release these suggestions as an inspectable changelog.

Finally, our coverage was affected by paywalls and text-mining agreements. We secured full text for \CorpusConverted{} of the \CorpusFiltered{} relevant papers, and the missing \CorpusNotObtained{} articles come disproportionately from specific publishers rather than a random slice of the literature. \CitationsUntraceable{}\% of chunk citations fail to link to real text blocks, mainly because the model copied example IDs straight from the prompt. While the extracted chemistry in these records is usually sound, confirming it requires checking the source paper by hand.

\section*{Conclusions}

We built a structured database of \ReleaseRecords{} homogeneous catalytic PET depolymerisation experiments extracted from \ReleasePapers{} papers, paired with an automated auditing pipeline to verify data quality. Rather than treating extracted values as final, an independent LLM judge read each source paper alongside its extracted records, accepting \ReleasePassRate{}\% of the database as written, proposing cited corrections for 30.5\%, and outright rejecting the remaining 2\%. Evaluated against manual review, the judge decisively outperformed a traditional heuristic baseline, acting as an effective triage filter that concentrates human attention on subtle errors in catalyst and solvent amounts.

Three practical lessons extend beyond this dataset. First, a single worked example provides all the guidance an extractor needs, saving tens of thousands of prompt tokens without sacrificing performance. Second, requiring chunk-level source citations anchors the model in the text, improving extraction accuracy while establishing verifiable provenance for every data point. Third, human verification budgets are spent most efficiently by focusing review exclusively on records where the automated graders disagree. 

This dataset provides a foundation for comparing homogeneous catalysts and modeling depolymerisation conditions. We release the prompts for LLM extraction and the LLM judge, the dataset, the judge's audit logs, the proposed changelogs, and an interactive review interface to support further data-driven research in polymer upcycling.

\section*{Author contributions}
\textbf{M.b.J.}: Conceptualisation, Investigation, Methodology, Supervision, Validation, Writing -- original draft. 
\textbf{K.K.B.H.}: Conceptualisation, Investigation, Methodology, Software, Supervision, Validation, Writing -- original draft. 
\textbf{E.B.}: Conceptualisation, Data curation, Investigation, Methodology, Software, Validation, Writing -- original draft. 
\textbf{A.M.}: Funding acquisition, Supervision, Writing -- review and editing. 
\textbf{M.G.}: Conceptualisation, Funding acquisition, Supervision, Writing -- review and editing.

\section*{Conflicts of interest}
There are no conflicts of interest to declare.

\section*{Data availability}
The dataset of \ReleaseRecords{} homogeneous catalytic PET depolymerisation experiments, comprising the extracted records, the LLM judge's verdicts and critiques, and its proposed corrections, is available at \url{https://github.com/Enkhnyam/pet-depolymerisation-database} under \textit{data/}, released under CC BY 4.0. The hand-curated reference key of 295 experiments across 24 articles used to validate the judge is provided under \textit{data/benchmark/}. The same repository contains the extraction and evaluation code, the prompts, and the manual review tooling, released under the MIT licence. The standalone, open-source software (\textit{Chemistry Data Extractor Toolkit}), including the interactive browser-based review interface, is available at \url{https://github.com/Enkhnyam/chemistry-data-extractor-toolkit}. Supplementary figures, extractor and judge prompts, additional discussion, and extended tables are provided in the Electronic Supplementary Information (ESI).

\section*{Acknowledgements}
We acknowledge support of the
Werner Siemens Foundation in the frame of the WSS Research Center ``catalaix''.

During the preparation of this study, the authors used large language models (OpenAI GPT-5.6 Sol and Anthropic Claude Opus 5) to assist with code development and refinement. All code implementations were reviewed, tested, and verified by the authors, who take full responsibility for the content of this work.

\balance

\bibliography{rsc}
\bibliographystyle{rsc}

\end{document}